\documentclass[dvips]{article}

\usepackage{fullpage}
\usepackage{pstricks}
\usepackage{amssymb}
\usepackage{amstext}
\usepackage{amsmath}
\usepackage{txfonts}
\usepackage{cmll}
\usepackage{txfonts}
\usepackage{rotating}
\usepackage{array}
\usepackage{pifont}

\usepackage{xy}
\xyoption{all}

\newdimen\proofrulebreadth \proofrulebreadth=.05em
\newdimen\proofdotseparation \proofdotseparation=1.25ex
\newdimen\proofrulebaseline \proofrulebaseline=2ex
\newcount\proofdotnumber \proofdotnumber=3
\let\then\relax
\def\hfi{\hskip0pt plus.0001fil}
\mathchardef\squigto="3A3B
\newif\ifinsideprooftree\insideprooftreefalse
\newif\ifonleftofproofrule\onleftofproofrulefalse
\newif\ifproofdots\proofdotsfalse
\newif\ifdoubleproof\doubleprooffalse
\let\wereinproofbit\relax
\newdimen\shortenproofleft
\newdimen\shortenproofright
\newdimen\proofbelowshift
\newbox\proofabove
\newbox\proofbelow
\newbox\proofrulename
\def\shiftproofbelow{\let\next\relax\afterassignment\setshiftproofbelow\dimen0 }
\def\shiftproofbelowneg{\def\next{\multiply\dimen0 by-1 }%
\afterassignment\setshiftproofbelow\dimen0 }
\def\setshiftproofbelow{\next\proofbelowshift=\dimen0 }
\def\setproofrulebreadth{\proofrulebreadth}

\def\prooftree{
\ifnum  \lastpenalty=1
\then   \unpenalty
\else   \onleftofproofrulefalse
\fi
\ifonleftofproofrule
\else   \ifinsideprooftree
        \then   \hskip.5em plus1fil
        \fi
\fi
\bgroup
\setbox\proofbelow=\hbox{}\setbox\proofrulename=\hbox{}%
\let\justifies\proofover\let\leadsto\proofoverdots\let\Justifies\proofoverdbl
\let\using\proofusing\let\[\prooftree
\ifinsideprooftree\let\]\endprooftree\fi
\proofdotsfalse\doubleprooffalse
\let\thickness\setproofrulebreadth
\let\shiftright\shiftproofbelow \let\shift\shiftproofbelow
\let\shiftleft\shiftproofbelowneg
\let\ifwasinsideprooftree\ifinsideprooftree
\insideprooftreetrue
\setbox\proofabove=\hbox\bgroup$\displaystyle 
\let\wereinproofbit\prooftree
\shortenproofleft=0pt \shortenproofright=0pt \proofbelowshift=0pt
\onleftofproofruletrue\penalty1
}

\def\eproofbit{
\ifx    \wereinproofbit\prooftree
\then   \ifcase \lastpenalty
        \then   \shortenproofright=0pt  
        \or     \unpenalty\hfil         
        \or     \unpenalty\unskip       
        \else   \shortenproofright=0pt  
        \fi
\fi
\global\dimen0=\shortenproofleft
\global\dimen1=\shortenproofright
\global\dimen2=\proofrulebreadth
\global\dimen3=\proofbelowshift
\global\dimen4=\proofdotseparation
\global\count255=\proofdotnumber
$\egroup  
\shortenproofleft=\dimen0
\shortenproofright=\dimen1
\proofrulebreadth=\dimen2
\proofbelowshift=\dimen3
\proofdotseparation=\dimen4
\proofdotnumber=\count255
}

\def\proofover{
\eproofbit 
\setbox\proofbelow=\hbox\bgroup 
\let\wereinproofbit\proofover
$\displaystyle
}%
\def\proofoverdbl{
\eproofbit 
\doubleprooftrue
\setbox\proofbelow=\hbox\bgroup 
\let\wereinproofbit\proofoverdbl
$\displaystyle
}%
\def\proofoverdots{
\eproofbit 
\proofdotstrue
\setbox\proofbelow=\hbox\bgroup 
\let\wereinproofbit\proofoverdots
$\displaystyle
}%
\def\proofusing{
\eproofbit 
\setbox\proofrulename=\hbox\bgroup 
\let\wereinproofbit\proofusing
\kern0.3em$
}

\def\endprooftree{
\eproofbit 
  \dimen5 =0pt
\dimen0=\wd\proofabove \advance\dimen0-\shortenproofleft
\advance\dimen0-\shortenproofright
\dimen1=.5\dimen0 \advance\dimen1-.5\wd\proofbelow
\dimen4=\dimen1
\advance\dimen1\proofbelowshift \advance\dimen4-\proofbelowshift
\ifdim  \dimen1<0pt
\then   \advance\shortenproofleft\dimen1
        \advance\dimen0-\dimen1
        \dimen1=0pt
        \ifdim  \shortenproofleft<0pt
        \then   \setbox\proofabove=\hbox{%
                        \kern-\shortenproofleft\unhbox\proofabove}%
                \shortenproofleft=0pt
        \fi
\fi
\ifdim  \dimen4<0pt
\then   \advance\shortenproofright\dimen4
        \advance\dimen0-\dimen4
        \dimen4=0pt
\fi
\ifdim  \shortenproofright<\wd\proofrulename
\then   \shortenproofright=\wd\proofrulename
\fi
\dimen2=\shortenproofleft \advance\dimen2 by\dimen1
\dimen3=\shortenproofright\advance\dimen3 by\dimen4
\ifproofdots
\then
        \dimen6=\shortenproofleft \advance\dimen6 .5\dimen0
        \setbox1=\vbox to\proofdotseparation{\vss\hbox{$\cdot$}\vss}%
        \setbox0=\hbox{%
                \advance\dimen6-.5\wd1
                \kern\dimen6
                $\vcenter to\proofdotnumber\proofdotseparation
                        {\leaders\box1\vfill}$%
                \unhbox\proofrulename}%
\else   \dimen6=\fontdimen22\the\textfont2 
        \dimen7=\dimen6
        \advance\dimen6by.5\proofrulebreadth
        \advance\dimen7by-.5\proofrulebreadth
        \setbox0=\hbox{%
                \kern\shortenproofleft
                \ifdoubleproof
                \then   \hbox to\dimen0{%
                        $\mathsurround0pt\mathord=\mkern-6mu%
                        \cleaders\hbox{$\mkern-2mu=\mkern-2mu$}\hfill
                        \mkern-6mu\mathord=$}%
                \else   \vrule height\dimen6 depth-\dimen7 width\dimen0
                \fi
                \unhbox\proofrulename}%
        \ht0=\dimen6 \dp0=-\dimen7
\fi
\let\doll\relax
\ifwasinsideprooftree
\then   \let\VBOX\vbox
\else   \ifmmode\else$\let\doll=$\fi
        \let\VBOX\vcenter
\fi
\VBOX   {\baselineskip\proofrulebaseline \lineskip.2ex
        \expandafter\lineskiplimit\ifproofdots0ex\else-0.6ex\fi
        \hbox   spread\dimen5   {\hfi\unhbox\proofabove\hfi}%
        \hbox{\box0}%
        \hbox   {\kern\dimen2 \box\proofbelow}}\doll%
\global\dimen2=\dimen2
\global\dimen3=\dimen3
\egroup 
\ifonleftofproofrule
\then   \shortenproofleft=\dimen2
\fi
\shortenproofright=\dimen3
\onleftofproofrulefalse
\ifinsideprooftree
\then   \hskip.5em plus 1fil \penalty2
\fi
}

\newcounter{countroman}
{\begin{list}{{\rm (\roman{countroman})}}{\usecounter{countroman}}}%
{\end{list}}

\newcounter{countalpha}
\newenvironment{anumerate}%
{\begin{list}{(\alph{countalpha})}{\usecounter{countalpha}}}%
{\end{list}}

\newcounter{countalphabf}
{\protect\begin{list}{{\rm (}{\bf \protect\alph{countalphabf}}{\rm%
)}}{\protect\usecounter{countalphabf}}}%
{\end{list}}

 \newcommand{\cuu}[1]{\Lbag{#1}\Rbag}  
  \newcommand{\upcpr}[1]{\cmn{#1}}  
  \newcommand{\docpr}[1]{\mnd{#1}}  
       \newcommand{\cpr}[1]{\lozenge{#1}} 

\newcommand{\fr}[2]{\mbox{$\frac{#1}{#2}$}}

\newcommand{\Prox}{{\sf Prox}}

\newcommand{\Matr}{{\sf Matr}}
\newcommand{\MMatr}{{\sf MMat}}
\newcommand{\CMatr}{{\sf CMat}}

\newcommand{\Id}{{\rm Id}}
\newcommand{\Leftt}{\Lambda}
\newcommand{\Rightt}{\Upsilon}
\newcommand{\Forgg}{{\sf W}}

\renewcommand{\to}{\rightarrow}

\newcommand{\tto}[1]{\xrightarrow{#1}}

\newcommand{\epi}{\twoheadrightarrow}
\newcommand{\ipe}{\twoheadleftarrow}

\newcommand{\inclusion}{\hookrightarrow}
\newcommand{\clusionin}{\hookleftarrow}
\newcommand{\bito}{\looparrowright}

\newcommand{\dis}[2]{\left({#1}\vdash {#2}\right)}
\newcommand{\dist}[3]{\left( {#2}\  {\vdash}\, {#3}\, \right)_{#1}}   

\newcommand{\mol}[3]{\left({#2}\,{\scriptstyle\models}\,{#3}\right)_{#1}}
\newcommand{\mo}[2]{\left({#1}\,{\models}\,{#2}\right)}

\newcommand{\WP}{\mbox{\Large $\wp$}}

\newcommand{\Pos}{{\sf Pos}}

\newcommand{\id}{{\rm id}}

\newcommand{\XXX}{{\cal X}}

\mathcode`\<="4268 
\mathcode`\>="5269 
\mathchardef\gt="313E 
\mathchardef\lt="313C 

 \def\pushright#1{{
    \parfillskip=0pt            
    \widowpenalty=10000         
    \displaywidowpenalty=10000  
    \finalhyphendemerits=0      
    \leavevmode                 
    \unskip                     
    \nobreak                    
    \hfil                       
    \penalty50                  
    \hskip.2em                  
    \null                       
    \hfill                      
    {#1}                        
    \par}}                      

 \def\qed{\pushright{$\square$}\penalty-700 \smallskip}

\newenvironment{prf}[1]{\begin{trivlist} \item[{\bf ~Proof}#1.]}%
{\qed\end{trivlist}}

\newcommand{\beq}{\begin{equation}}
\newcommand{\eeq}{\end{equation}}
\newcommand{\ba}[1]{\begin{array}{#1}}
\newcommand{\ea}{\end{array}}
\newcommand{\bea}{\begin{eqnarray}}
\newcommand{\eea}{\end{eqnarray}}
\newcommand{\bear}{\begin{eqnarray*}}
\newcommand{\eear}{\end{eqnarray*}}
\newcommand{\bpr}{\begin{prf}{}}
\newcommand{\epr}{\end{prf}}
\newcommand{\bprf}[1]{\begin{prf}{#1}}
\newcommand{\eprf}{\end{prf}}

\newtheorem{proposition}{Proposition}[section]
\newtheorem{thrm}[proposition]{Theorem}
\newtheorem{corr}[proposition]{Corollary}
\newtheorem{lemm}[proposition]{Lemma}
\newtheorem{defn}[proposition]{Definition}

\newcommand{\op}[1]{\overline{#1}}
\newcommand{\dual}[1]{{#1}^\ddag}
\newcommand{\Up}[1]{\ \Uparrow\negthinspace{#1}}
\newcommand{\Do}[1]{\ \Downarrow\negthinspace{#1}}
\newcommand{\UD}[1]{\ \Updownarrow\negthinspace{#1}}

\newcommand{\dow}[1]{\overleftarrow{#1}}

\newcommand{\up}[1]{\overrightarrow{#1}}

\newcommand{\cmn}{{\rm \Delta}}
\newcommand{\mnd}{\nabla}

\newcommand{\tprod}{\textstyle\prod}
\newcommand{\tcoprod}{\textstyle\coprod}
\newcommand{\ohm}{{\rm \Omega}}

\title{Quantitative Concept Analysis}

\author{Dusko Pavlovic\\
Royal Holloway, University of London, and University of Twente\\
{\small Email:~dusko.pavlovic@rhul.ac.uk}}

\date{}

\begin{document}
\maketitle

\begin{abstract}
Formal Concept Analysis (FCA) begins from a context, given as a binary relation between some objects and some attributes, and derives a lattice of concepts, where each concept is given as a set of objects and a set of attributes, such that the first set consists of all objects that  satisfy all attributes in the second, and vice versa. Many applications, though, provide contexts with quantitative information, telling not just whether an object satisfies an attribute, but also quantifying this satisfaction. 
Contexts in this form arise as rating matrices in recommender systems, as occurrence matrices in text analysis, as pixel intensity matrices in digital image processing, etc. Such applications have attracted a lot of attention, and several numeric extensions of FCA have been proposed. We propose the framework of \emph{proximity sets (proxets)}, which subsume partially ordered sets (posets) as well as metric spaces. One feature of this approach is that it extracts from quantified contexts  quantified concepts, and thus allows full use of the available information. Another feature is that the categorical approach  allows analyzing any universal properties that the classical FCA and the new versions may have, and thus provides structural guidance for aligning and combining the approaches.
\end{abstract}

\section{Introduction}\label{Introduction}
Suppose that the users $U = \left\{\mbox{Abby}, \mbox{Dusko}, \mbox{Stef}, \mbox{Temra}, \mbox{Luka}\right\}$ provide the following star ratings for the items $J = \left\{\mbox{"Nemo", "Crash" , "Ikiru", "Bladerunner"}\right\}$
\begin{center}
\begin{tabular}{|m{1cm}||m{2cm}|m{2cm}|m{2cm}|m{2cm}|}
\hline
 & "Nemo" & "Crash" & "Ikiru" & "Bladerunner"\\
\hline \hline 
Abby & $\star\star\star\, \star$ & $\star\star\star\star\star$ & $\star\, \star$ & $\star\star\star\, \star$  \\
\hline
Dusko & $\star\, \star$ & $\star\, \star$ &$\star\star\star\, \star$& $\star\star\star\star\star$ \\
\hline
Stef & $\star\, \star$ & $\star\star\star\star\star$ &$\star\star\, \star$ & $\star\, \star$ \\
 \hline
 Temra & $\star$ & $\star\star\star$  & $\star\star\star$ & $\star\star\star\, \star$ \\
 \hline
Luka & $\star\star\star\star\star$ & $\star$ & $\star$ &  $\star\, \star$ \\
\hline
\end{tabular}
\end{center}
This matrix $\Phi = (\Phi_{iu})_{J\times U}$ contains some information about the relations between these users' tastes, and about the relations between the styles of the items (in this case movies) that they rated. The task of data analysis is to extract that information. In particular, given a \emph{context} matrix $\Phi:J\times U\to R$ like in the above table
, the task of \emph{concept} analysis is to detect, on one hand, the latent concepts of \emph{taste}, shared by some of the users in $U$, and on the other hand the latent concepts of \emph{style}, shared by some of the items in $J$. In Formal Concept Analysis (FCA) \cite{WilleR:FCA,FCA-book,Carpineto-Romano:book,FCA-foundations,Poelmans:survey}, the latent concepts are expressed as sets: a taste $t$ is a set of users, i.e. a map $U\tto t \{0,1\}$, whereas a style $s$ is a set of items, i.e. a map $J\tto s \{0,1\}$. We explore a slightly refined notion of concept, which tells not just whether two users (resp. two items) share the same taste (resp. style) or not, but it also quantifies the degree of \emph{proximity} of their tastes (resp. styles). This is formalized by expressing a taste as a map $U\tto \tau [0,1]$, and a style as a map $J\tto \sigma [0,1]$. The value $\tau_u$ is thus a number from the interval [0,1], telling how close is the taste $\tau$ to the user $u$; whereas the value $\sigma_i$ tells how close is the item $i$ to the style $\sigma$. These concepts are \emph{latent}, in the sense that they are not given in advance, but mined from the context matrix, just like in FCA, and similarly like in Latent Semantic Analysis (LSA) \cite{LSA}. Although the extracted concepts are interpreted differently for the users in $U$ and for the items in $J$ (i.e. as the tastes and the styles, respectively) it turns out that the two obtained concept structures are isomorphic, just like in FCA and LSA.  However, our approach allows initializing a concept analysis session by some prior concept structures, which allow building upon the results of previous analyses, from other data sets, or specified by the analyst. This allows introducing different conceptual backgrounds for the users in  $U$ and for the items in $J$.


\paragraph{Related work and background.} The task of capturing quantitative data in FCA was recognized early on. The simplest approach is to preprocess any given numeric data into relational contexts  by introducing thresholds, and then apply the standard FCA method  \cite{WilleR:scaling,FCA-book}. This basic approach has been extended in several directions, e.g. Triadic Concept Analysis \cite{Kaytoue:biclusters,KAYTOUE:2010,lehmann1995triadic} and Pattern Structures \cite{ganter2001pattern,Kaytoue:revisiting,KaytoueM:gene}, and refined for many application domains. A different way to introduce numeric data into FCA is to allow \emph{fuzzy} contexts, as binary relations evaluated in an abstract lattice of truth values $L$. The different ways to lift the FCA constructions along the inclusion $\{0,1\}\hookrightarrow L$ have led to an entire gamut of different versions of fuzzy FCA \cite{BelohlavekR:book,BelohlavekR:APAL,burusco1998construction,burusco2008study,krajci2005generalized}, surveyed in \cite{belohlavek2005whatis}. With one notable exception, all versions of fuzzy FCA input quantitative data in the form as fuzzy relations, and output qualitative concept lattices in the standard form. The fact that numeric input data are reduced to the usual lattice outputs can be viewed as an advantage, since the outputs can then be presented, and interpreted, using the available FCA visualization tools and methods. On the other hand, only a limited amount of information contained in a numeric data set can be effectively captured in lattice displays. The practices of spectral methods of concept analysis \cite{Azar,LSA,KorenY:factorization}, pervasive in web commerce, show that the quantitative information received in the input contexts can often be preserved in the output concepts, and effectively used in ongoing analyses. Our work has been motivated by the idea that suitably refined FCA constructions could output concept structures with useful quantitative information, akin to the concept eigenspaces of LSA.  It turns out that the steps towards quantitative concepts on the FCA side have previously been made by B{\v{e}}lohl{\'a}vek in \cite{BelohlavekR:APAL}, where fuzzy concept lattices derived from fuzzy contexts were proposed and analyzed. This is the mentioned notable exception from the other fuzzy and quantitative approaches to FCA, which all derive just qualitative concept lattices from quantitative contexts. B{\v{e}}lohl{\'a}vek's basic definitions turn out to be remarkably close to the definitions we start from in the present paper, in spite of the fact that his goal is to generalize FCA using carefully chosen fuzzy structures, whereas we use enriched categories with the ultimate goal to align FCA with the spectral methods for concept analysis, such as LSA. Does this confirm that the structures obtained in both cases naturally arise from the shared FCA foundations, rather than from either the fuzzy or the categorical approach? The ensuing analyses, however, shed light on these structures from essentially different angles, and open up completmentary views: while B{\v{e}}lohl{\'a}vek provides a detailed analysis of the internal structure of fuzzy concept lattices, we provide a high level view of their universal properties, from which some internal features follow, and which offers guidance through the maze of the available structural choices. Combining the two methods seems to open interesting alleys for future work.
 
Our motivating example suggests that our goals might be related to those of \cite{FCA-collab}, where an  FCA approach to recommender systems was proposed. However, the authors of \cite{FCA-collab} use FCA to tackle the problem of \emph{partial} information (the missing ratings) in recommender systems, and they abstract away the \emph{quantitative} information (contained in the available ratings); whereas our goal is to capture this quantitative information, and we leave the problem of partial information aside for the moment.

\paragraph{Outline of the paper.} In Sec.~\ref{Proxets-sec} we introduce \emph{proximity sets (proxets)}, the mathematical formalism supporting the proposed generalization of FCA.  Some constructions and notations used throughout the paper are introduced in Sec.~\ref{notations-sec}. Since proxets generalize posets, in Sec.~\ref{Vectors-sec} we introduce the corresponding generalizations of infimum and supremum, and spell out the  basic completion constructions, and the main properties of the infimum (resp. supremum) preserving morphisms. In Sec.~\ref{Matrices-sec}, we study context matrices over proximity sets, and describe their decomposition, with a universal property analogous to the Singular Value Decomposition of matrices in linear algebra. Restricting this decomposition from proxets to discrete posets (i.e. sets) yields FCA. The drawback of this quantitative version of FCA is that in it a finite context generally allows an infinite proxet of concepts, whereas in the standard version of FCA, of course, finite contexts lead to finite concept lattices. This problem is tackled in Sec.~\ref{Basic-sec}, where we show how the  users and the items, as related in the context, induce a finite generating set of concepts. Sec.~\ref{Conclusions-sec} provides a discussion of the obtained results and ideas for the future work. 

\section{Proxets}\label{Proxets-sec}
\subsection{Definition, intuition, examples}\label{intuitions}
\paragraph{Notation.} Throughout the paper, the order and lattice structure of the interval $[0,1]$ are denoted by $\leq$, $\wedge$ and $\vee$, whereas $\cdot$ denotes the multiplication in it.

\begin{defn}\label{def-proximity}
A \emph{proximity} over a set $A$ is a map $\dis{}{} :A\times A\to [0,1]$ which is
\begin{itemize}
\item reflexive: $\dis x x = 1$,
\item transitive: $\dis x y \cdot \dis y z\ \leq\ \dis x z$, and
\item antisymmetric: $\dis x y = 1 = \dis y x \ \Longrightarrow\  x=y$
\end{itemize}
If only reflexity and transitivity are satisfied, and not antisymmetry, then we have an \emph{intensional} proximity map. The antisymmetry condition is sometimes called \emph{extensionality}. A(n intensional) \emph{proximity set}, or \emph{proxet}, is a set equipped with a(n intensional) proximity map. A \emph{proximity} (or \emph{monotone}) morphism between the proxets $A$ and $B$ is a function $f:A\to B$ such that all $x,y\in A$ satisfy $\dist A x y   \leq  \dist B {fx} {fy}$.
We denote by $\Prox$ the category of proxets and their morphisms.
\end{defn}

\paragraph{Categorical view.} A categorically minded reader can understand intensional proxets as categories  enriched \cite{KellyGM:book-enriched} over the poset $[0,1]$ viewed as a category, with the monoidal structure induced by the multiplication. In the presence of reflexivity and transitivity, $\dis x y = 1$ is equivalent with $\forall z.\ \dis z x \leq \dis z y$, and with $\forall z.\ \dis x z \geq \dis y z$. A proximity map is thus asymmetric if and only if $\left(\forall z.\ \dis z x = \dis z y\right)\Rightarrow x=y$, and if and only if $\left(\forall z.\ \dis z x = \dis z y\right)\Rightarrow x=y$. This means that extensional proxets correspond to \emph{skeletal} $[0,1]$-enriched categories.

\subsubsection{Examples.} The first example of a proxet is the interval $[0,1]$ itself, with the proximity
\bea\label{vdash}
\dist {[0,1]} x y & = & \begin{cases} \frac{y}{x} & \mbox{ if } y\lt x\\
1 & \mbox{ otherwise}
\end{cases}
\eea
Note that $\dis {} {} : [0,1]\times [0,1]\to [0,1]$ is now an \emph{operation} on [0,1], satisfying
\bea\label{MP}
(x\cdot y) \leq z & \iff & x \leq \dis y z
\eea

A wide family of examples follows from the fact that proximity sets (proxets) generalize partially ordered sets (posets), in the sense that any poset $S$ can be viewed as a proxet $\Forgg S$, with the proximity induced by the partial ordering $\underset S \sqsubseteq$ as follows: 
\bea\label{pos}
\dist {\Forgg S} x y & = &\begin{cases} 1 & \mbox{ if } x\underset S \sqsubseteq y\\
0 & \mbox{ otherwise}
\end{cases}
\eea 
The proxet $\Forgg S$ is intensional if and only if $S$ is just a \emph{preorder}, in the sense that the relation $\underset S \sqsubseteq$ is just transitive and reflexive. The other way around, any (intensional) proxet $A$ induces two posets (resp. preorders), $\Rightt A$ and $\Leftt A$, with the same underlying set and 
\[x\underset{\Rightt A}\sqsubseteq y \iff  \dist A x y = 1 \qquad \qquad \qquad x\underset{\Leftt A}\sqsubseteq y  \iff \dist A x y \gt 0  
\]
Since the constructions $\Forgg$, $\Rightt$ and $\Leftt$, extended on maps, preserve monotonicity, a categorically minded reader can easily confirm that we have three functors, which happen to form two adjunctions $\Leftt \dashv \Forgg \dashv \Rightt : \Prox \to \Pos$. Since $\Forgg:\Pos \inclusion \Prox$ is an embedding, $\Pos$ is thus a reflective and correflective subcategory of $\Prox$. This means that  $\Leftt \Forgg S = S =  \Rightt \Forgg S$ holds for every poset $S$, so that posets are exactly the proxets where the proximities are evaluated only in 0 or 1; and that $\Leftt A$ and  $\Rightt A$ are respectively the initial and the final poset induced by the proxet $A$, as witnessed by the obvious morphisms $\Forgg\Rightt A \to A \to \Forgg \Leftt A$. The same universal properties extend to a correspondence between intensional proxets and preorders.

A different family of examples is induced by metric spaces: any metric space $X$ with a distance  map $d:X\times X\to [0,\infty]$ can be viewed as a proxet with the proximity  map
\bea\label{metric}
\dis x y & = & 2^{-d(x,y)}
\eea
Proxets are thus a common generalization of posets and metric spaces. But the usual metric distances are symmetric, i.e. satisfy $d(x,y) = d(y,x)$, whereas the proximities need not be. The inverse of \eqref{metric} maps any proximity to a \emph{quasi}-metric $d(x,y) = -\log \dis x y$ \cite{Wilson:quasi-metric}, whereas intensional proximities induce \emph{pseudo}-quasi-metrics \cite{Kim:pseudo-quasi-metric}.
For a concrete family of examples of quasi-metrics, take any family of sets $\XXX \subseteq \WP X$, and define
\bear
d(x,y) & = & | y \setminus x |
\eear
The distance of $x$ and $y$ is thus the number of elements of $y$ that are not in $x$. This induces the proximity $\dis x y = 2^{-|y\setminus x|}$. If $\XXX$ is a set of documents, viewed as bags (multisets) of terms, then both constructions can be generalized to count the difference in the numbers of the occurrences of terms in documents, and the set difference becomes multiset subtraction.

\paragraph{Proximity or distance?} The isomorphism $-\log x: [0,1] \rightleftarrows [0,\infty]:2^{-x}$ is easily seen to lift to an isomorphism between the category of proxets, as categories enriched over the multiplicative monoid $[0,1]$ and the category of generalized metric spaces, as categories enriched over the additive monoid $[0,\infty]$. Categorical studies of generalized metric spaces were initiated in  \cite{LawvereFW:metric}, continued in denotational semantics of programming languages \cite{WagnerK,RuttenJ:metric,Schellekens:Yoneda}, and have recently turned out to be useful for quantitative distinctions in ecology \cite{LeinsterT:species}. The technical results of this paper could equivalently be stated in the framework of generalized metric spaces. While this would have an advantage of familiarity to certain communities, the geometric intuitions that come with metrics turn out to be misleading when imposed on the applications that are of interest here. The lifting of infima and suprema is fairly easy from posets to proxets, but leads to mysterious looking operations over metrics. In any case, the universal properties of matrix decompositions do not seem to have been studied in either framework so far.

\subsection{Derived proxets and notations}\label{notations-sec}
Any proxets $A, B$ give rise to other proxets by following standard constructions:
\begin{itemize}
\item the \emph{dual} (or \emph{opposite}) proxet $\op A$, with the same underlying set and the proximity $\dist {\op A} x y = \dist A y x$;
\item the \emph{product} proxet $A\times B$ over the cartesian product of the underlying sets, and the proximity $\dist{A\times B} {x, u} {y, v}= \dist A x y \wedge \dist B u v$

\item the \emph{power} proxet $B^A$ over the monotone maps, i.e. $\Prox(A, B)$ as the underlying set, with the proximity $\dist {B^A} f g = \bigwedge_{x\in A} \dist B {fx} {gx}$.
\end{itemize}
There are natural correspondences of proxet morphisms
\[\Prox(A, B)\times \Prox(A,C)  \cong  \Prox(A, B\times C)\quad\mbox{and} \quad 
\Prox(A\times B, C) \cong  \Prox(A, C^B)
\]
\paragraph{Notations.} In any proxet $A$, it is often convenient to abbreviate $\dist A x y = 1$ to $x \underset A \leq y$. For $f,g:A\to B$, it is easy to see that $f\underset{B^A} \leq g$ if and only if $fx\underset B \leq gx$ for all $x\in A$.

\section{Vectors, limits, adjunctions}\label{Vectors-sec}
\subsection{Upper and lower vectors}
Having generalized posets to proxets, we proceed to lift the concepts of the least upper bound and the greatest lower bound. Let $(S,\sqsubseteq)$ be a poset and let $L, U\subseteq S$ be a lower set and an upper set, respectively, in the sense that 
\[(x\sqsubseteq y\ \mbox{and}\ y \in L) \Rightarrow x\in L\qquad \qquad (x\in U \ \mbox{and}\ x \sqsubseteq y) \Rightarrow y\in U\]
Then an element denoted $\bigsqcup L$ is supremum of $L$, and $\bigsqcap U$ is the infimum of $U$, if all $x, y \in A$ satisfy
\bea
\textstyle \bigsqcup L\leq y & \iff & \forall x.\ \left(x\in L \Rightarrow x \sqsubseteq y\right)\label{join}\\
x\leq \textstyle \bigsqcap U & \iff & \forall y.\ \left(y\in U\Rightarrow x\sqsubseteq y\right)\label{meet}
\eea
We generalize these definitions  
 to proxet limits in (\ref{pjoin}-\ref{pmeet}). To generalize the lower sets, over which the suprema are taken, and the upper sets for infima, observe that any upper set $U\subseteq S$ corresponds to a monotone map $\up U:S\to \{0,1\}$, whereas every lower set $L$ corresponds to an antitone map $\dow L:\op S \to \{0,1\}$, where $\op S$ is the dual proxet defined in Sec.~\ref{notations-sec}.

\begin{defn} \label{infsup} An  \emph{upper} and a \emph{lower vector} in a proxet $A$ are the monotone maps $\up \upsilon : A\to [0,1]$ and $\dow \lambda: \op A \to [0,1]$. The sets of vectors $\Up A = \op{[0,1]^{A}}$ and $\Do A = [0,1]^{\op A}$ form proxets, with the proximities  computed in terms of the infima in $[0,1]$, as
\[ \dist {\Up A} {\up \upsilon} {\up \tau} = \bigwedge_{x\in A} \dist A {\up \tau_x}{\up \upsilon_x}\qquad \mbox{ and } \qquad \dist {\Do A} {\dow \lambda} {\dow \mu } = \bigwedge_{x\in A} \dist A {\dow \lambda_x}{\dow \mu_x}\]
\end{defn}

\paragraph{Remark.} Note that the defining condition for upper vectors $\dis x y \leq \dis {\up \upsilon_x} {\up \upsilon_y}$, and the defining condition for lower vectors $\dis x y \leq \dis {\dow \lambda_x} {\dow \lambda_y}$ are respectively equivalent with
\[
\up \upsilon_x\cdot \dis x y \leq \up \upsilon_y \qquad\mbox{ and } \qquad \dis x y \cdot \dow \lambda_y \leq \dow \lambda_x
\]

\subsection{Limits}\label{Limits-sec}
\begin{defn}
The \emph{upper limit} or \emph{supremum} $\tcoprod \dow \lambda$ of the lower vector $\dow \lambda$ and the \emph{lower limit} or \emph{infimum} $\tprod \up\upsilon$ of the upper vector $\up \upsilon$ are the elements of $A$ that satisfy for every $x,y\in A$
\bea
\dist A {\tcoprod \dow \lambda} y & = & \bigwedge_{x\in A} \dow \lambda_x \vdash \dist A x y\label{pjoin}\\
\dist A x {\tprod \up \upsilon}  & = & \bigwedge_{y\in A} \up \upsilon_y \vdash \dist A x y\label{pmeet}
\eea
The proxet $A$ is \emph{complete} under infima (resp. suprema) if every upper (resp. lower) vector has an infimum (resp. supremum), which thus yield the operations $\tprod  :\ \ \Up A \to A$ and $\tcoprod:\ \ \Do A \to A$
\end{defn}

\paragraph{Remarks.} Condition \eqref{pjoin} generalizes \eqref{join}, whereas \eqref{pmeet} generalizes \eqref{meet}. Note how proximity operation $\vdash$ over [0,1], defined in \eqref{vdash}, plays in (\ref{pjoin}--\ref{pmeet}) the role that the implication $\Rightarrow$ over $\{0,1\}$ played in (\ref{join}--\ref{meet}). This is justified by the fact that $\vdash$ is adjoint to the multiplication in $[0,1]$, in the sense of \eqref{MP}, in the same sense in which $\Rightarrow$ is adjoint to the meet in $\{0,1\}$, or in any Heyting algebra, in the sense of $(x \wedge y) \leq z \iff x\leq (y\Rightarrow z)$. 

An element $w$ of a poset $S$ is an upper bound of $L\subseteq S$ if it satisfies just one direction of \eqref{join}, i.e. 
\bear
(w\sqsubseteq y) & \Longrightarrow &  \forall x.\ \left(x\in L \Rightarrow x \sqsubseteq y\right)
\eear 
Ditto for the lower bounds. In a proxet $A$, $u$ is an upper bound of $\dow \lambda$ and $\ell$ is a lower bound of $\up \upsilon$ if all $x,y\in A$ satisfy
\[
\dist A {u} y \leq  \bigwedge_{x\in A} \dow \lambda_x \vdash \dist A x y\qquad \mbox{ and } \qquad
\dist A x {\ell}   \leq  \bigwedge_{y\in A} \up \upsilon_y \vdash \dist A x y
\]
Using \eqref{MP} and instantiating $y$ to $u$ in the first inequality, and $x$ to $\ell$ in the second one, these conditions can be shown to be equivalent with $\dow \lambda_x  \leq  \dist A x u$ and $\up \upsilon_y \leq \dist A \ell y$,
which characterize the upper and the lower bounds in proxets. 

\subsection{Completions}
Each element $a$ of a proxet $A$ induces two \emph{representable} vectors
\begin{alignat*}{4}
{\rm \cmn} a\ :\ A&\to  [0,1] & \qquad \qquad && \mnd a\ :\ \op A&\to  [0,1]\\
x & \mapsto \dist A a x  & && x &\mapsto \dist A x a
\end{alignat*}
It is easy to see that these maps induce proximity morphisms $\cmn : A\to \Up A$ and $\mnd : A \to \Do A$,
which correspond to the categorical \emph{Yoneda embeddings} \cite[Sec.~III.2]{MacLane:CWM}. They make $\Up A$ into the lower completion, and $\Do A$ into the upper completion of the proxet $A$.

\begin{proposition}\label{completion}
$\Up A$ is upper complete and $\Do A$ is lower complete. Moreover, they are universal, in the sense that
\begin{itemize}
\item any monotone $f:A\to C$ into a complete proxet $C$ induces a unique $\prod$-preserving morphism $f_\#: \Up A\to C$ such that $f = f_\#\circ \cmn$;
\item any monotone $g:A\to D$ into a cocomplete proxet $D$ induces a unique $\coprod$-preserving morphism $g^\#: \Do A\to D$ such that $g = g^\#\circ \mnd$.
\end{itemize}
\[
\xymatrix@-1pc{
&&& \Up A \ar@{-->}[dd]^{\exists ! f_\#}\\
A\ar[urrr]^{\cmn} \ar[drrr]_{\forall f} \\
&&& C
}\qquad \qquad 
\xymatrix@-1pc{
&&& \Do A \ar@{-->}[dd]^{\exists ! g^\#}\\
A\ar[urrr]^{\mnd} \ar[drrr]_{\forall g} \\
&&& D
}
\] 
\end{proposition}

\subsection{Adjunctions}
\begin{proposition}\label{adjunctions}
For any proximity morphism $f:A\to B$ holds $(a)\iff(b)\iff(c)$ and $(d)\iff(e)\iff(f)$, 
where
\begin{anumerate}
\item $f\left(\tcoprod \dow \lambda\right) = \tcoprod f\left(\dow \lambda\right)$
\item $\exists f_\ast:B\rightarrow A\ \forall x\in A\ \forall y\in B.\  \dist B {fx} y = \dist A x {f_\ast y}$
\item $\exists f_\ast:B\rightarrow A.\  \id_A \leq {f_\ast f} \wedge {ff_\ast} \leq \id_B$
\item $f\left(\tprod \up \upsilon\right) = \tprod f\left(\up \upsilon\right)$
\item $\exists f^\ast : B\rightarrow A\ \forall x\in A\ \forall y\in B.\  \dist B {f^\ast y} x = \dist A y {fx}$
\item $\exists f^\ast:B\rightarrow A.\  {f^\ast f} \leq \id_A \wedge \id_B \leq {ff^\ast}$
\end{anumerate}
The morphisms $f^\ast$ and $f_\ast$ are unique, whenever they exist.
\end{proposition}

\begin{defn}
An \emph{upper adjoint} is a proximity morphism satisfying (a-c) of Prop.~\ref{adjunctions}; a \emph{lower adjoint} satisfies (d-f).
A \emph{(proximity) adjunction} between proxets $A$ and $B$ is a pair of proximity morphisms 
$f^\ast: A\rightleftarrows B: f_\ast$
 related as in (b-c) and (e-f). 
\end{defn}

%
%
%

%
%
%
%

\subsection{Projectors and nuclei}
\begin{proposition}\label{projections}
For any adjunction $f^\ast: A\rightleftarrows B: f_\ast$ holds $(a)\iff(b)$ and $(c)\iff(d)$,  
where
\begin{anumerate}
\item $\forall xy\in B.\ \dist A {f_\ast x} {f_\ast y} = \dist B x y$
\item $f^\ast f_\ast = \id_B$
\item $\forall xy\in A.\ \dist B {f^\ast x} {f^\ast y} = \dist A x y$
\item $f_\ast f^\ast  = \id_A$
\end{anumerate}
\end{proposition}

\begin{defn}
An adjunction satisfying (a-b)  of Prop.~\ref{projections} is an \emph{upper projector}; an adjunction satisfying (c-d) is a \emph{lower projector}. The upper (resp. lower) component of an upper (resp. lower) projector is called  the upper (lower) \emph{projection}. The other component (i.e. the one in (a), resp. (c)) is called the upper (lower) \emph{embedding}.
\end{defn}

\begin{proposition}\label{factorization}
Any upper (lower) adjoint factors, uniquely up to isomorphism, through an upper (lower) projection followed by an upper (lower) embedding through the proxet
\bear
\cuu f & = & \left\{<x,y>\in A\times B\ |\  f^\ast x = y \wedge x = f_\ast y\right\}
\eear
%
\end{proposition}


\begin{defn}
A \emph{nucleus} of the adjunction $f^\ast: A\rightleftarrows B: f_\ast$ consists of a proxet $\cuu f$ together with
\begin{itemize}
\item embeddings $A \stackrel {e_\ast} \clusionin \cuu f \stackrel{e^\ast} \inclusion B$
\item projections $A \stackrel {p^\ast} \epi \cuu f \stackrel{p_\ast} \ipe B$
\end{itemize}
such that $f^\ast = e^\ast p^\ast$ and $f_\ast = e_\ast p_\ast$. 
\end{defn}

\subsection{Cones and cuts}
The \emph{cone operations} are the proximity morphisms $\cmn^\#$ and $\mnd_\#$
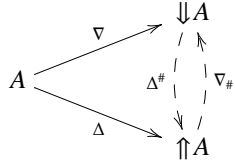
\[
\xymatrix@-1pc{
&&& \Do A \ar@/_/@{-->}[dd]_{\cmn^\#}\\
A\ar[urrr]^{\mnd} \ar[drrr]_{\cmn} \\
&&& \Up A \ar@/_/@{-->}[uu]_{\mnd_\#}
}
\] 
These morphisms are induced by the universal properties of the Yoneda embeddings $\mnd$ and  $\cmn$ as completions, stated in Prop.~\ref{completion}. Since by definition $\cmn^\#$ preserves suprema, and $\mnd_\#$ preserves infima, Prop.~\ref{adjunctions} implied that each of them is an adjoint, and it is not hard to see that they form the adjunction $\cmn^\# : \Do A \rightleftarrows \Up A: \mnd_\#$. Spelling them out yields
\[
\left(\cmn^\# \dow \lambda\right)_a = \bigwedge_{x\in A} \dow \lambda_x \vdash \dis x a \qquad \qquad \left(\mnd_\# \up \upsilon \right)_a = \bigwedge_{x\in A} \up \upsilon_x \vdash \dis a x
\]
Intuitively, $\left(\cmn^\# \dow \lambda\right)_a$ is the proximity of $\dow \lambda$ to $a$ as its upper bound, as discussed in Sec.~\ref{Limits-sec}. Visually, $\left(\cmn^\# \dow \lambda\right)_a$ thus measures the \emph{cone} from $\dow \lambda$ to $a$, whereas $\left(\mnd_\# \up \upsilon \right)a$ measures the cone from $a$ to $\up \upsilon$.

\begin{proposition}\label{saturated-prop}
For every $\dow \lambda \in \Do A$ every $\up \upsilon \in \Up A$ holds
\begin{alignat*}{5}
\dow \lambda  
& \leq \mnd_\# \cmn^\# \dow\lambda
&\qquad\mbox{and}\qquad& 
\dow \lambda \geq \mnd_\# \cmn^\# \dow\lambda
&\ \iff \exists \up\upsilon.\ \dow \lambda = \cmn^\#\up \upsilon\\
\up \upsilon  & \leq \cmn^\# \mnd_\# \up \upsilon
&\qquad\mbox{and}\qquad&
\up \upsilon \geq \cmn^\# \mnd_\#  \up \upsilon
&\ \iff \exists \dow \lambda.\ \up \upsilon = \mnd_\#\dow \lambda
\end{alignat*}
The transpositions make the following subproxets isomorphic
\bear
\left(\Do A\right)_{\mnd_\# \cmn^\#} & = & \left\{\dow \lambda\in \Do A\ |\ \dow \lambda  = \mnd_\# \cmn^\# \dow\lambda\right\}\\
\left(\Up A\right)_{\cmn^\#\mnd_\# } & = & \left\{\up\upsilon \in \Up A\ |\ \up \upsilon = \cmn^\# \mnd_\#  \up \upsilon\right\}
\eear
\end{proposition}



\begin{defn}
The vectors in $\left(\Do A\right)_{\mnd_\# \cmn^\#}$ and $\left(\Up A\right)_{\cmn^\#\mnd_\# }$ are called \emph{cones}. The associated cones $\dow \gamma \in \left(\Do A\right)_{\mnd_\# \cmn^\#}$ and $\up \gamma \in \left(\Up A\right)_{\cmn^\#\mnd_\# }$ such that $\dow \gamma = \mnd_\# \up \gamma$ and $\up \gamma = \cmn^\# \dow \gamma$ a \emph{cut} $\gamma = <\dow \gamma, \up \gamma>$ in proxet $A$. Cuts form a proxet $\UD A$, isomorphic with $\left(\Do A\right)_{\mnd_\# \cmn^\#}$ and $\left(\Up A\right)_{\cmn^\#\mnd_\# }$, with the proximity
\[
\dist {\UD A} \gamma \varphi \ =\ \dist {\Do A} {\dow \gamma} {\dow \varphi} \ =\ \dist {\Up A} {\up \gamma} {\up \varphi}
\]
\end{defn}


\begin{lemm}
The $\UD A$-infima are constructed in $\Do A$, and $\UD A$ suprema are constructed in $\Up A$. 
\end{lemm}


\begin{corr}
A proxet $A$ has all suprema if and only if it has all infima. \end{corr}

%

\paragraph{Dedekind-MacNeille completion is a special case.} If $A$ is a poset, viewed by \eqref{pos} as the proxet $\Forgg A$, then $\UD \Forgg A$ is the Dedekind-MacNeille completion of $A$ \cite{MacNeille}. The above construction extends the Dedekind-MacNeille completion to the more general framework of proxets, in the sense that it satisfies in the universal property of the Dedekind-MacNeille completion \cite{Banaschewski-Bruns}. The construction seems to be novel in the familiar frameworks of metric and quasi-metric spaces. However, Quantitative Concept Analysis requires that we lift this construction to matrices.

\section{Proximity matrices and their decomposition}\label{Matrices-sec}

\subsection{Definitions, connections}
\begin{defn}\label{matrix-def}
A {\em proximity matrix} $\Phi$ from proxet $A$ to proxet $B$ is a vector $\Phi: \op A \times B\to [0,1]$. We write it as $\Phi:A\bito B$, and write its value $\Phi(x,y)$ at $x\in A$ and $y\in B$ in the form $\mol \Phi x y$. The matrix composition of $\Phi:A\bito B$ and $\Psi: B\bito C$ is defined
 \bear
 \mol{(\Phi\, ; \Psi)} x z & = & \bigvee_{y\in B} \mol \Phi x y \cdot \mol \Psi y z 
 \eear 
With this composition and the identity matrices $\Id_A:A\times A\to [0,1]$ where $\Id_A(x,x') \ = \ \dist A x {x'}$, proxets and proxet matrices form the category $\Matr$.
 \end{defn}
 
\paragraph{Remark.} Note that the defining condition $\dis u x \cdot \dis y v \leq \dis {\mol \Phi x y} {\mol \Phi u v}$, which says that $\Phi$ is a proximity morphism $\op A \times B\to [0,1]$, can be equivalently written
\bea\label{matrix-cond}
\dis u x \cdot \mol \Phi x y \cdot \dis y v \ \leq \ \mol \Phi u v
\eea

\begin{defn} 
The \emph{dual} $\dual \Phi:B\bito A$ of a matrix $\Phi:A\bito B$ has the entries
\bear
\mol {\dual \Phi} y x & = & \bigwedge_{\substack{u\in A\\ v\in B}} \mol \Phi u v \vdash \left( \dist A u x \cdot  \dist B y v \right)
\eear
A matrix $\Phi:A\bito B$ where $\Phi^{\ddag\ddag} = \Phi$ is called a \emph{suspension}.
\end{defn}

\paragraph{Remarks.} 
It is easy to see by Prop.~\ref{saturated-prop} that $\dist \Phi x y \leq \dist {\Phi^{\ddag\ddag}} x y$ holds for all $x\in A$ and $y\in B$, and that $\Phi$ is a suspension if and only if there is some $\Psi: B\bito A$ such that $\Phi = \Psi^\ddag$. It is easy to see that $\Phi \leq \Psi\Rightarrow\dual \Phi \geq \dual \Psi$, and thus $\Phi \leq \Phi^{\ddag\ddag}$ implies $\dual \Phi = \Phi^{\ddag\ddag\ddag}$.

\begin{defn}
The matrices $\Phi:A\bito B$ and $\Psi:B\bito A$ form a \emph{connection} if\\ $\Phi\, ;\Psi \leq \Id_A$ and $\Psi\, ;\Phi \leq \Id_B$.
\end{defn}

\begin{proposition}\label{Galois-prop}
$\Phi:A\bito B$ and $\Phi^\ddag:B\bito A$ always form a connection.
\end{proposition}


\begin{defn}\label{density-def}
A matrix $\Phi: A\bito B$ is  \emph{embedding} if $\Phi\, ; \dual \Phi = \Id_A$; and a \emph{projection} if $\dual \Phi\, ; \Phi = \Id_B$.
\end{defn}

\begin{defn}\label{decomposition-def}
A \emph{decomposition} of a matrix $\Phi:A\bito B$ consists of a proxet $D$, with
\begin{itemize}
\item projection matrix $P: A\bito D$, i.e. $\dist D d {d'} = \bigvee_{x\in A} \mol {\dual P} d x \cdot \mol P x {d'}$,
\item embedding matrix $E: D\bito B$, i.e. $\dist D d {d'}  =  \bigvee_{y\in B} \mol {E} d y \cdot \mol {\dual E} y {d'}$,
\end{itemize}
such that $\Phi = P\, ; E$, i.e. $\mol \Phi x y  =  \bigvee_{d\in D} \mol P x d \cdot \mol E d y$.\end{defn}

%
%
%
%
%
%
%

\paragraph{Matrices as adjunctions.} A matrix $\Phi : A\bito B$ can be equivalently presented as either of the proximity morphisms $\Phi_\bullet$ and $\Phi^\bullet$, which extend to $\Phi_\ast$ and $\Phi^\ast$ using Thm.~\ref{completion}
\[
\prooftree
\prooftree
\op A\times B\tto \Phi [0,1]
\justifies
A \tto {\Phi_\bullet} \Up B\qquad \qquad B \tto {\Phi^\bullet} \Do A
\endprooftree
\justifies
\Do A \tto {\Phi^\ast} \Up B\qquad \qquad \Up B \tto {\Phi_\ast} \Do A
\endprooftree
\] 
\vspace{-1\baselineskip}
\bea\label{adj}
\left(\Phi^\ast\dow \lambda\right)_b  =  \bigwedge_{x\in A} \dow \lambda_x \vdash \mol \Phi x b\qquad\qquad
\left(\Phi_\ast\up \upsilon\right)_a =  \bigwedge_{y\in B} \up \upsilon_y \vdash \mol \Phi a y
\eea
Both extensions, and their nucleus, are summarized in diagram \eqref{decomp}.
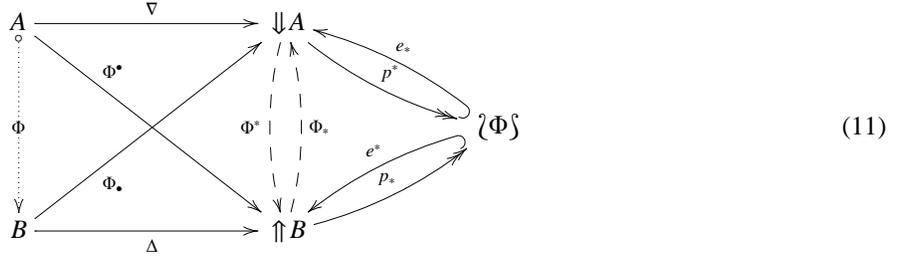
 \begin{equation}\label{decomp}
 \begin{split}
\xymatrix
{
A \ar[rrr]^{\mnd} \ar@{o.>}[dd]|\Phi \ar[ddrrr]^(.3){\Phi^\bullet} &&& \Do A \ar@/_/@{-->}[dd]_{\Phi^\ast}
\\ &&& 
&& 
\ \ \cuu \Phi 
\ar@/_/@{^{(}->}[ull]_{e_\ast} 
\ar@/_/@{^{(}->}[dll]_{e^\ast}
\ar@/^/@{<<-}[ull]_{p^\ast} 
\ar@/^/@{<<-}[dll]_{p_\ast}
\\
B \ar[rrr]_\cmn \ar[uurrr]_(.3){\Phi_\bullet}&&& \Up B \ar@/_/@{-->}[uu]_{\Phi_\ast}
}
\end{split}
\end{equation}
%
The adjunction $\Phi^\ast:\Do A\rightleftarrows \Up B:\Phi_\ast$ means that
\[
\dist {\Up B} {\Phi^\ast \dow \lambda} {\up \upsilon}\ =\ \bigwedge_{y\in B}  {\up \upsilon_y} \vdash (\Phi^\ast \dow \lambda)_y \ = \ \ \bigwedge_{x\in A}  {\dow \lambda_x} \vdash (\Phi_\ast \up \upsilon)_x\ =\  \dist {\Do A} {\dow \lambda} {\Phi_\ast \up \upsilon}
\]
holds. The other way around, it can be shown that any adjunction between $\Do A$ and $\Up B$ is completely determined by the induced matrix from $A$ to $B$. 

\begin{proposition}
The matrices $\Phi\in \Matr(A,B)$ are in a bijective correspondence with the adjunctions $\Phi^\ast: \Do A\rightleftarrows \Up B : \Phi_\ast$.
\end{proposition}


\subsection{Matrix decomposition through nucleus}
Prop.~\ref{saturated-prop} readily lifts to matrices.

\begin{proposition}\label{matrix-saturated-prop}
For every $\dow \alpha \in \Do A$ every $\up \beta \in \Up B$ holds
\begin{alignat*}{5}
\dow \alpha  
& \leq \Phi_\ast \Phi^\ast \dow\alpha
&\qquad\mbox{and}\qquad& 
\dow \alpha \geq \Phi_\ast \Phi^\ast \dow\alpha
&\ \iff \exists \up\beta\in \Up B.\ \dow \alpha = \Phi^\ast\up \beta\\
\up \beta  & \leq \Phi^\ast \Phi_\ast \up \beta
&\qquad\mbox{and}\qquad&
\up \beta \geq \Phi^\ast \Phi_\ast  \up \beta
&\ \iff \exists \dow \alpha \in \Do A.\ \up \beta = \Phi_\ast\dow \alpha
\end{alignat*}
The adjunction $\Phi^\ast:A\rightleftarrows B:\Phi_\ast$ induces the isomorphisms between the following proxets
\bear
\cuu \Phi_A & = & \left\{\dow \alpha\in \Do A\ |\ \dow \alpha  = \Phi_\ast \Phi^\ast \dow\alpha\right\}\\
\cuu \Phi_B & = & \left\{\up\beta \in \Up B\ |\ \up \beta = \Phi^\ast \Phi_\ast  \up \beta\right\}\\
\cuu \Phi & = & \left\{\gamma = <\dow \gamma, \up \gamma> \in \Do A\ \times\! \Up B\ |\ \dow \gamma  = \Phi_\ast \up \gamma \wedge \Phi^\ast \dow \gamma = \up\gamma\right\}
\eear
with the proximity
\[
\dist {\cuu\Phi} \gamma \varphi \ =\ \dist {\Do A} {\dow \gamma} {\dow \varphi} \ =\ \dist {\Up B} {\up \gamma} {\up \varphi}
\]
\end{proposition}

\begin{defn}
$\cuu \Phi$ is called the \emph{nucleus} of the matrix $\Phi$. Its elements are the \emph{$\Phi$-cuts}. 
\end{defn}

\begin{thrm}\label{complete-thm}
The matrix $\Phi: A\bito B$ decomposes through $\cuu \Phi$ into
\begin{itemize}
\item the projection  $P^\ast:A \bito \cuu \Phi$ with $\mol {P^\ast} x {<\dow \alpha, \up \beta>} = \dow \alpha_x$, and
\item the embedding $E^\ast:\cuu \Phi \bito B$ with $\mol {E^\ast} {<\dow \alpha, \up \beta>} y = \up \beta_y$ 
\end{itemize}
\end{thrm}

\subsection{Universal properties}\label{FCA-univ}
Any proxet morphism $f:A\to B$ induces two matrices, $\ohm f :A\bito B$ and $\mho f : B\bito A$ with
\[
\mol {\ohm f} x y = \dist B {fx} y \qquad \qquad \mol{\mho f} y x = \dist B y {fx}
\]

%
%
%
%

\begin{defn}
A \emph{proximity matrix morphism} from a matrix $\Phi: F_0\bito F_1$ to $\Gamma: G_0\bito G_1$ consists of pair of monotone maps $h_0:F_0\to G_0$ and $h_1:F_1\to G_1$ such that 
\begin{itemize}
\item $\ohm h_0\, ; \Gamma = \Phi\, ; \ohm h_1$,
\item $h_0$ preserves any $\tcoprod$ that may exist in $F_0$,
\item $h_1$ preserves any $\tprod$ that may exist in $F_1$.
\end{itemize}
Let $\MMatr$ denote the category of proxet matrices and matrix morphisms. Let $\CMatr$ 
 denote the full subcategory spanned by proximity matrices between complete proxets.
\end{defn}

\begin{proposition}\label{universal}
$\CMatr$ is reflective in $\MMatr$ along $\cuu{-}:\MMatr \rightleftarrows \CMatr:U$
\end{proposition}


\paragraph{Posets and FCA.} If $A$ and $B$ are posets, a $\{0,1\}$-valued proxet matrix $\Phi:A\bito B$ can be viewed as a subposet $\Phi\subseteq A\times B$, lower closed in $A$ and upper closed in $B$.  The adjunction $\Phi^\ast:A\rightleftarrows B:\Phi_\ast$ is the Galois connection induced by $\Phi$, and the posetal nucleus $\cuu \Phi$ is now the complete lattice such that
\begin{itemize}
\item $A\tto\mnd \Do A \epi \cuu \Phi$ is $\vee$-generating and $\wedge$-preserving,
\item $B\tto\mnd \Up B \epi \cuu \Phi$ is $\wedge$-generating and $\vee$-preserving.
\end{itemize}
When $A$ and $B$ are discrete posets, i.e. with all elements incomparable, then any binary relation $R\subseteq A\times B$ can be viewed as a proxet matrix between them. Restricting to the vectors that take their values in 0 and 1 yields $\Do A \cong (\WP A, \subseteq)$ and $\Up B \cong (\WP B, \supseteq)$. The concept lattice of FCA then arises from the Galois connection $R^\ast: \Do A \rightleftarrows \Up B:R_\ast$ 
 as the concept lattice $\cuu R$. Restricted to $\{0,1\}$-valued matrices between discrete sets $A$ and $B$, Prop.~\ref{universal} thus yields a universal construction of a lattice $\vee$-generated by $A$ and $\wedge$-generated by $B$. The FCA concept lattice derived from a context $\Phi$ is thus its posetal nucleus $\cuu \Phi$. This universal property is closely related with the methods and results of \cite{Banaschewski-Bruns,Gehrke06}. 
 
 \paragraph{Lifting The Basic Theorem of FCA.} The Basic Theorem of FCA says that every complete lattice can be realized as a concept lattice, namely the the one induced by the context of its own partial order. For quantitative concept analysis, this is an immediate consequence of Prop.\ref{universal}, which implies a proxet $A$ is complete if and only if $\Id_A = \cuu{\Id_A}$. Intuitively, this just says that nucleus, as a completion, preserves the structure that it completes, and must therefore be idempotent, as familiar from the Dedekind-MacNeille construction. It should be noted that this property does not generalize beyond proxets. 

\section{Representable concepts and their proximities}\label{Basic-sec}
\subsection{Decomposition without completion}
The problem with factoring matrices $\Phi:A\bito B$ through $\cuu \Phi$ in practice is that $\cuu \Phi$ is a large, always infinite structure. The proxet $\cuu \Phi$ is the completion of the matrix $\Phi:A\bito B$ in the sense that it is
\begin{itemize}
\item the subproxet of the $\tcoprod$-completion $\Do A$ of $A$, spanned by the vectors $\dow \alpha = \Phi_\ast \Phi^\ast \dow \alpha$,
\item the subproxet of the $\tprod$-completion $\Up B$ of $B$, spanned by the vectors $\up \beta = \Phi^\ast \Phi_\ast \up\beta$.
\end{itemize}
Since there  are always uncountably many lower and upper vectors, and the completions $\Do A$ and $\Up B$ are infinite, $\cuu \Phi$ follows suit. But can we extract a small set of generators of $\cuu \Phi$, still supporting a decomposition of the matrix $\Phi$.

\begin{defn}\label{basis}
The \emph{representable concepts} induced $\Phi$ are the elements of the completion $\cuu \Phi$ induced the representable vectors, i.e.
\begin{itemize}
\item \emph{lower representable concepts} $\docpr\Phi  = \left\{\left<\Phi_\ast \Phi^\ast \mnd a, \Phi^\ast \mnd a\right>\ |\ a\in A\right\}$
\item \emph{upper representable concepts} $\upcpr\Phi = \left\{ \left<\Phi_\ast \cmn b, \Phi^\ast \Phi_\ast \cmn b\right>\ |\ b\in B\right\}$
\item \emph{representable concepts} $ \cpr\Phi  = \docpr \Phi \cup \upcpr \Phi$
\end{itemize}
\end{defn}

\paragraph{Notation.} The elements of $\cpr \Phi$ are written in the form $\cpr x = <\dow {\cpr x}, \up{\cpr x}>$, and thus
\begin{align*}
\dow{\cpr a} & = \Phi_\ast \Phi^\ast \mnd a & \up{\cpr a} & = \Phi^\ast \mnd a\\
\dow{\cpr b} & = \Phi_\ast \cmn b & \up{\cpr b} & = \Phi^\ast \Phi_\ast \cmn b
\end{align*}


\begin{thrm}\label{decomp-thm}
For any proxet matrix $\Phi: A\bito B$, the restriction  of the decomposition $A\stackrel {P^\ast} \bito \cuu \Phi \stackrel {E^\ast} \bito B$ from Thm.~\ref{complete-thm} along the inclusion $\cpr \Phi \inclusion \cuu \Phi$ to the representable concepts yields a decomposition  $A\stackrel {P} \bito \cpr \Phi \stackrel {E} \bito B$ which still satisfies Def.~\ref{decomposition-def}. More precisely, the matrices
\begin{itemize}
\item $P : \op A \times {\cpr \Phi} \inclusion \op A \times \cuu \Phi \tto {P^\ast} [0,1]$ 
\item $E: \op{\cpr \Phi} \times B \inclusion \op{\cuu \Phi} \times B \tto {E^\ast} [0,1]$
\end{itemize}
are such that  $P: A \bito \cpr \Phi$ is a projection, $E: \cpr \Phi \bito B$ is an embedding, and $P\, ; E = \Phi$.
\end{thrm}

%
%

\subsection{Computing proximities of representable concepts}\label{computing}
To apply these constructions to the ratings matrix from Sec.~\ref{Introduction}, we first express the star ratings as numbers between 0 and 1. 
\begin{center}
\begin{tabular}[c]{|c||c|c|c|c|}
\hline
 &\hspace{.3em} \it n \hspace{.3em}  & \hspace{.3em}\it c\hspace{.3em} &\hspace{.3em} \it i\hspace{.3em} &\hspace{.3em} \it b \hspace{.3em} \\
\hline \hline 
\hspace{.3em} \it a\hspace{.3em} & \fr 4 5 & 1 & \fr 2 5  & \fr 4 5   \\[.5ex]
\hline
 \it d & \fr 2 5  & \fr 2 5  & \fr 4 5   & 1 \\[.5ex]
\hline
 \it s & \fr 2 5  & 1 &\fr 3 5  & \fr 2 5  \\[.5ex]
 \hline
  \it t &\fr 1 5  & \fr 3 5   & \fr 3 5  & \fr 4 5  \\[.5ex]
 \hline
 \it l & 1 & \fr 1 5  & \fr 1 5  &  \fr 2 5  \\[.5ex]
\hline
\end{tabular}
\end{center}
where we also abbreviated the user names to $U = \{A,D,S,T,L\}$ and the item names to $J=\{n,c,i,b\}$.
Now we can compute the representable concepts $\cpr \varphi\in \cpr \Phi$ according to Def.~\ref{basis}, using \eqref{adj}:
{\small
\begin{align*}
(\dow{\cpr j})_u & = \left( \bigwedge_{\ell \in J} (\cmn j)_\ell \vdash \mo u \ell\right) =  \mo u j &\quad (\up{\cpr j})_k & = \left(\bigwedge_{x\in U} (\up{\cpr j})_x \vdash \mo x k\right) =\left( \bigwedge_{x\in U} \mo x j \vdash \mo x k\right)\\
(\up{\cpr u})_j & = \left(\bigwedge_{x\in U} (\mnd u)_x \vdash \mo x j\right)= \mo u j &\quad (\dow{\cpr u})_v & = \left(\bigwedge_{\ell \in J} (\dow{\cpr u})_\ell \vdash \mo v \ell\right) =\left( \bigwedge_{\ell \in J} \mo u \ell \vdash \mo v \ell\right)
\end{align*}
}
\hspace{-1ex}Since $\dow{\cpr\varphi} = \Phi_\ast \up{\cpr \varphi}$ and $\Phi^\ast \dow{\cpr \varphi} = \up{\cpr \varphi}$, it suffices to compute one component of each pair $\cpr \varphi = <\dow {\cpr \varphi}, \up{\cpr \varphi}>$, say the first one. So we get
\begin{alignat*}{5}
\dow{\cpr n}  &= \begin{pmatrix} \fr 4 5 & \fr 2 5 & \fr 2 5 & \fr 1 5 & 1
\end{pmatrix} 
&\qquad\qquad& 
\dow{\cpr c} & = \begin{pmatrix}1 & \fr 2 5 & 1 & \fr 3 5 & \fr 1 5
\end{pmatrix}
&\qquad\qquad&
\dow{\cpr \imath} & = \begin{pmatrix}\fr 2 5 & \fr 4 5 & \fr 3 5 & \fr 3 5 & \fr 1 5
\end{pmatrix}  
\\
\dow{\cpr b} & = \begin{pmatrix}\fr 4 5 & 1 & \fr 2 5 & \fr 4 5 & \fr 1 5
\end{pmatrix} 
&&
\dow{\cpr a}  &= \begin{pmatrix} 1 & \fr 2 5  & \fr 1 2 & \fr 1 4 & \fr 1 5 \end{pmatrix}
&&
\dow{\cpr d} & = \begin{pmatrix} \fr 1 2  & 1  & \fr 2 5 & \fr 1 2 & \fr 1 4 \end{pmatrix} 
\\
\dow{\cpr s} & = \begin{pmatrix} \fr 2 3  & \fr 2 5  & 1 & \fr 1 2 & \fr 1 5 \end{pmatrix}
&&
\dow{\cpr t} & = \begin{pmatrix} \fr 2 3  & \fr 2 3  & \fr 1 2 & 1 & \fr 1 3 \end{pmatrix}
&&
\dow{\cpr l} & = \begin{pmatrix} \fr 4 5  & \fr 2 5  & \fr 2 5 & \fr 1 5 &1 \end{pmatrix}
\end{alignat*}
The proximities between all representable concepts can now be computed in the form 
\bear
\dist {\cpr \Phi} x y\ =\ \dist {\cpr \Phi} {\cpr x} {\cpr y} & = & \bigwedge_{u\in U} \dow{\cpr x}_u \vdash \dow{\cpr y}_u
\eear
since the proximity in $\cpr \Phi$ is just the proximity in $\mnd \Phi$, which is a subproxet ot $\Do U$, so its proximity is  by Def.~\ref{infsup} the pointwise minimum. Hence 
\begin{center}
\begin{tabular}[c]{|c||c|c|c|c|c|c|c|c|c|}
\hline
$\vdash$ &\hspace{.3em} \it n \hspace{.3em}  & \hspace{.3em}\it c\hspace{.3em} &\hspace{.3em} \it i\hspace{.3em} &\hspace{.3em} \it b \hspace{.3em} &\hspace{.3em} \it a \hspace{.3em} &\hspace{.3em} \it d \hspace{.3em} &\hspace{.3em} \it s \hspace{.3em} &\hspace{.3em} \it t \hspace{.3em} &\hspace{.3em} \it l \hspace{.3em}  \\
\hline \hline 
\hspace{.3em} \it n\hspace{.3em} & 1 & \fr 1 5 & \fr 1 5  & \fr 1 5 & \fr 1 5 & \fr 1 4 & \fr 1 5 & \fr 1 3 & 1   \\[.5ex]
\hline
\hspace{.3em} \it c\hspace{.3em} & \fr 1 3 & 1 & \fr 2 5  & \fr 2 5 & \fr 5 {12} & \fr 2 5 & \fr 2 3 & \fr 1 2 & \fr 1 3  
 \\[.5ex]
\hline
\hspace{.3em} \it i \hspace{.3em} & \fr 1 3 & \fr 1 2 & 1  & \fr 2 3 & \fr 5 {12} & \fr 2 3 & \fr 1 2 & \fr 5 6 & \fr 1 3   \\[.5ex]
\hline
\hspace{.3em} \it b\hspace{.3em} & \fr 1 4 & \fr 2 5 & \fr 1 2  & 1 & \fr 5 {16} & \fr 5 8 & \fr 2 5 & \fr 2 3 & \fr 1 4   \\[.5ex]
\hline
\hspace{.3em} \it a\hspace{.3em} & \fr 4 5 & 1 & \fr 2 5  & \fr 4 5 & 1 & \fr 1 2 & \fr 2 3 & \fr 2 3 & \fr 4 5  \\[.5ex]
\hline
 \it d & \fr 2 5  & \fr 2 5  & \fr 4 5   & 1 & \fr 2 5 & 1 & \fr 2 5 & \fr 2 3 & \fr 2 5 \\[.5ex]
\hline
 \it s & \fr 2 5  & 1 &\fr 3 5  & \fr 2 5 & \fr 1 2 & \fr 2 5 & 1 & \fr 1 2 & \fr 2 5  \\[.5ex]
 \hline
  \it t &\fr 1 5  & \fr 3 5   & \fr 3 5  & \fr 4 5 & \fr 1 4 & \fr 1 2 & \fr 1 2 & 1 & \fr 1 5 \\[.5ex]
 \hline
 \it l & 1 & \fr 1 5  & \fr 1 5  &  \fr 2 5 & \fr 1 5 & \fr 1 4 & \fr 1 5 & \fr 1 3 & 1 \\[.5ex]
\hline
\end{tabular}
\end{center}
The bottom five rows of this table display the values of the representable concepts themselves
\bea
\dist{\cpr \Phi} u j\ &  = &\ \mol \Phi u j \label{first}\\
 \dist{\cpr\Phi} u v\ & = & \ \bigwedge_{\ell \in J} {\mol \Phi v \ell} \vdash {\mol \Phi u \ell}\label{second}
\eea
for $u,v \in U$ and $j\in J$, because $\dist {\cpr \Phi}{\dow{\cpr u}} {\dow{\cpr x}} = \dow{\cpr x}_u$ follows from the general fact that $\dist{\Do A}{\mnd a} {\dow \lambda} = \dow\lambda_a$. The upper four rows display the values
\bea\label{third}
\dist{\cpr \Phi} j k\ & = & \ \bigwedge_{x \in U} {\mol \Phi x j} \vdash {\mol \Phi x k} \\
 \dist{\cpr\Phi} j u\ & = & \ \bigwedge_{x \in U} \mol \Phi x j \vdash \dist{\cpr \Phi} x u \ \ =\ \ \bigwedge_{\ell \in J} \mol{\Phi} u \ell \vdash \dist {\cpr \Phi} j \ell\label{fourth}
\eea
for $u\in U$ and $j,k \in J$. Intuitively, these equations can be interpreted as follows: 
\begin{itemize}
\item \eqref{second} \emph{the proximity $\dis u v$ measures how well $\mo v \ell$ approximates $\mo u \ell$}:
\begin{itemize}
\item $u$'s liking $\mo u \ell$ of any movie $\ell$  is at least $\dis u v \cdot \mo v \ell$. 
\end{itemize}
\item \eqref{third} \emph{the proximity $\dis j k$ measures how well $\mo x j$ approximates $\mo x k$}
\begin{itemize}
\item any user $x$'s rating $\mo x k$ is at least $\mo x j \cdot \dis j k$,
\end{itemize}
\item \eqref{fourth} \emph{the proximity $\dis j u$ measures how well $j$'s style approximates $u$'s taste}
\begin{itemize}
\item any $x$'s proximity $\dis x u$ to $u$ is at least $\mo x j \cdot \dis j u$,
\item $j$'s proximity $\dis j \ell$ to any $\ell$ is at least $\dis j u \cdot \mo u \ell$.
\end{itemize}
\end{itemize}
Since $\dis a l = \fr 4 5$, it would make sense for Abby to accept Luka's recommendations, but not the other way around, since $\dis l a = \frac 1 5$. Although Temra's rating of "Ikiru" is just $\dis t i = \fr 3 5$, "Ikiru" is a good test of her taste, since her rating of it is close to both Dusko's and Stefan's ratings.

\paragraph{Latent concepts?} While the proximities between each pair of users and items, i.e. between the induced \emph{representable} concepts, provide an interesting new view on their relations, the task of determining the \emph{latent} concepts remains ahead. What are the dominant tastes around which the users coalesce? What are the dominant styles that connect the items? What will such concepts look like? Formally, a dominant concept is a highly biased cut: in a high proximity of some of the representable concepts, and distant from the others. One way to find such cuts is to define the concepts of \emph{cohesion} and \emph{adhesion} of a cut along the lines of \cite{PavlovicD:CSR08}, and solve the corresponding optimization problems. Although there is no space to expand the idea in the present paper, some of the latent concepts can be recognized already by inspection of the above proximity table (recalling that each cut is both a supremum of users' and an infimum of items' representations).

\section{Discussion and future work}\label{Conclusions-sec}
What has been achieved? We generalized posets to proxets in Sec.~\ref{Proxets-sec} and \ref{Vectors-sec}, and lifted in Sec.~\ref{Matrices-sec} the FCA concept lattice construction to the corresponding construction over proxets, that allow capturing quantitative information. Both constructions share the same universal property, captured by the nucleus functor in Sec.~\ref{FCA-univ}. In both cases, the concepts are captured by \emph{cuts}, echoing Dedekind's construction of the reals, and MacNeille's minimal completion of a poset. But while finite contexts yield finite concept lattices in FCA, in our analysis they yield infinitely many quantitative concepts.  This is a consequence of introducing the infinite set of quantities [0,1]. The same phenomenon occurs in LSA \cite{LSA}, which allows the entire real line of quantities, and the finite sets of users and items span real vector spaces, that play the same role as our proxet completions. The good news is that the infinite vector space of latent concepts in LSA comes with a canonical basis of finitely many singular vectors, and that our proxet of latent concepts also has a finite generator, spelled out in Sec.~\ref{Basic-sec}. The bad news is that the generator described there is not a canonical basis of dominant latent concepts, with the suitable extremal properties, but an ad hoc basis determined by the given sets of users and items. Due to a lack of space, the final step of the analysis, finding the basis of dominant latent concepts, had to be left for a future paper. This task can be reduced to some familiar optimization problems. 

More interestingly, and perhaps more effectively, this task can also addressed using qualitative FCA and its concept scaling methods \cite{WilleR:scaling}. The most effective form of concept analysis may thus very well be a combination of quantitative and qualitative analysis tools. Our analysis of the numeric matrix, extracted from the given star ratings, should be supplemented by standard FCA analyses of a family of relational contexts scaled by various thresholds. We conjecture that the resulting relational concepts will be the projections of the dominant latent concepts arising from quantitative analysis. If that is the case, then the relational concepts can be used to guide computation of quantitative concepts.

This view of the quantitative and the qualitative concept analyses as parts of a putative general FCA toolkit raises an interesting question of their relation with LSA and the spectral methods of concept analysis \cite{LSA,Azar}, which seem different. Some preliminary discussions on this question can be found in \cite{PavlovicD:QI08,PavlovicD:FAST10}. While FCA captures a \emph{particle} view of network traffic, where the shortest path determines the proximity of two network nodes, LSA corresponds to the \emph{wave} view of the traffic, where the proximity increases with the number of paths. Different application domains seem to justify different views, and call for a broad view of all concept mining methods as parts of the same general toolkit.

\paragraph{Acknowledgements.} Anonymous reviewers' suggestions helped me to improve the paper, and to overcome some of my initial ignorance about the FCA literature. I am particularly grateful to Dmitry Ignatov, who steered the reviewing process with a remarkable patience and tact. I hope that my work will justify the enlightened support, that I encountered in these first contacts with the FCA community.


\begin{thebibliography}{10}

\bibitem{Azar}
Yossi Azar, Amos Fiat, Anna Karlin, Frank McSherry, and Jared Saia.
\newblock Spectral analysis of data.
\newblock In {\em Proceedings of the thirty-third annual {ACM Symposium on
  Theory of Computing}}, STOC '01, pages 619--626, New York, NY, USA, 2001.
  ACM.

\bibitem{Banaschewski-Bruns}
B.~Banaschewski and G.~Bruns.
\newblock {Categorical characterization of the MacNeille completion}.
\newblock {\em Archiv der Mathematik}, 18(4):369--377, September 1967.

\bibitem{BelohlavekR:book}
R.~B{\v{e}}lohl{\'a}vek.
\newblock {\em Fuzzy relational systems: foundations and principles},
  volume~20.
\newblock Plenum Publishers, 2002.

\bibitem{BelohlavekR:APAL}
R.~B{\v{e}}lohl{\'a}vek.
\newblock Concept lattices and order in fuzzy logic.
\newblock {\em Annals Pure Appl. Logic}, 128(1-3):277--298, 2004.

\bibitem{belohlavek2005whatis}
R.~Belohl{\'a}vek and V.~Vychodil.
\newblock What is a fuzzy concept lattice?
\newblock In et~al Sergei O.~Kuznetsov, editor, {\em Proceedings of RSFDGrC
  2011}, volume 6743 of {\em Lecture Notes in Computer Science}, pages 19--26.
  Springer, 2011.

\bibitem{RuttenJ:metric}
M.~M. Bonsangue, F.~van Breugel, and J.~J. M.~M. Rutten.
\newblock Generalized metric spaces: completion, topology, and power domains
  via the yoneda embedding.
\newblock {\em Theor. Comput. Sci.}, 193(1-2):1--51, 1998.

\bibitem{burusco1998construction}
A.~Burusco and R.~Fuentes-Gonz{\'a}lez.
\newblock {Construction of the L-fuzzy concept lattice}.
\newblock {\em Fuzzy Sets and systems}, 97(1):109--114, 1998.

\bibitem{burusco2008study}
A.~Burusco and R.~Fuentes-Gonz{\'a}lez.
\newblock {The study of the L-fuzzy concept lattice}.
\newblock {\em Mathware \& Soft Computing}, 1(3):209--218, 2008.

\bibitem{Carpineto-Romano:book}
Claudio Carpineto and Giovanni Romano.
\newblock {\em Concept Data Analysis: Theory and Applications}.
\newblock John Wiley \& Sons, 2004.

\bibitem{LSA}
Scott~C. Deerwester, Susan~T. Dumais, Thomas~K. Landauer, George~W. Furnas, and
  Richard~A. Harshman.
\newblock Indexing by {Latent Semantic Analysis}.
\newblock {\em Journal of the American Society of Information Science},
  41(6):391--407, 1990.

\bibitem{FCA-collab}
Patrick du~Boucher-Ryan and Derek~G. Bridge.
\newblock Collaborative recommending using {Formal Concept Analysis}.
\newblock {\em Knowl.-Based Syst.}, 19(5):309--315, 2006.

\bibitem{ganter2001pattern}
B.~Ganter and S.~Kuznetsov.
\newblock Pattern structures and their projections.
\newblock {\em Conceptual Structures: Broadening the Base}, pages 129--142,
  2001.

\bibitem{WilleR:scaling}
B.~Ganter and R.~Wille.
\newblock Conceptual scaling.
\newblock {\em Institute for Mathematics and Its Applications}, 17:139, 1989.

\bibitem{FCA-foundations}
Bernhard Ganter, Gerd Stumme, and Rudolf Wille, editors.
\newblock {\em Formal Concept Analysis, Foundations and Applications}, volume
  3626 of {\em Lecture Notes in Computer Science}. Springer, 2005.

\bibitem{FCA-book}
Bernhard Ganter and Rudolf Wille.
\newblock {\em Formal Concept Analysis: Mathematical Foundations}.
\newblock Springer, Berlin/Heidelberg, 1999.

\bibitem{Gehrke06}
Mai Gehrke.
\newblock Generalized kripke frames.
\newblock {\em Studia Logica}, 84(2):241--275, 2006.

\bibitem{Kaytoue:biclusters}
Mehdi Kaytoue, Sergei~O. Kuznetsov, Juraj Macko, Wagner~Meira Jr., and Amedeo
  Napoli.
\newblock Mining biclusters of similar values with triadic concept analysis.
\newblock In {\em Proceedings of CLA 2011}. CLA, 2011.

\bibitem{KAYTOUE:2010}
Mehdi Kaytoue, Sergei~O. Kuznetsov, and Amedeo Napoli.
\newblock Pattern mining in numerical data: {Extracting} closed patterns and
  their generators.
\newblock Research Report RR-7416, INRIA, October 2010.

\bibitem{Kaytoue:revisiting}
Mehdi Kaytoue, Sergei~O. Kuznetsov, and Amedeo Napoli.
\newblock Revisiting numerical pattern mining with formal concept analysis.
\newblock In {\em Proceedings of IJCAI 2011}, pages 1342--1347. AAAI, 2011.

\bibitem{KaytoueM:gene}
Mehdi Kaytoue, Sergei~O. Kuznetsov, Amedeo Napoli, and S\'e bastien Duplessis.
\newblock {Mining gene expression data with pattern structures in formal
  concept analysis}.
\newblock {\em Inf. Sci.}, 10(181):1989--2001, 2011.

\bibitem{KellyGM:book-enriched}
Gregory~Maxwell Kelly.
\newblock {\em Basic Concepts of Enriched Category Theory}.
\newblock Number~64 in London Mathematical Society Lecture Note Series.
  Cambridge University Press, 1982.
\newblock Reprinted in Theory and Applications of Categories, No. 10 (2005) pp.
  1-136.

\bibitem{KorenY:factorization}
Yehuda Koren, Robert~M. Bell, and Chris Volinsky.
\newblock Matrix factorization techniques for recommender systems.
\newblock {\em IEEE Computer}, 42(8):30--37, 2009.

\bibitem{krajci2005generalized}
S.~Kraj{\v{c}}i.
\newblock A generalized concept lattice.
\newblock {\em Logic Journal of IGPL}, 13(5):543--550, 2005.

\bibitem{Schellekens:Yoneda}
H.~P. K\"{u}nzi and M.~P. Schellekens.
\newblock On the yoneda completion of a quasi-metric space.
\newblock {\em Theor. Comput. Sci.}, 278(1-2):159--194, 2002.

\bibitem{LawvereFW:metric}
F.~William Lawvere.
\newblock Metric spaces, generalised logic, and closed categories.
\newblock {\em Rendiconti del Seminario Matematico e Fisico di Milano},
  43:135--166, 1973.

\bibitem{lehmann1995triadic}
F.~Lehmann and R.~Wille.
\newblock A triadic approach to formal concept analysis.
\newblock {\em Conceptual structures: applications, implementation and theory},
  pages 32--43, 1995.

\bibitem{LeinsterT:species}
Tom Leinster and Christina Cobbold.
\newblock {Measuring diversity: the importance of species similarity}.
\newblock {\em Ecology}, 2012.
\newblock to appear.

\bibitem{MacNeille}
Holbrook~Mann MacNeille.
\newblock Extensions of partially ordered sets.
\newblock {\em Proc. Nat. Acad. Sci.}, 22(1):45--50, 1936.

\bibitem{MacLane:CWM}
Saunders {Mac\thinspace Lane}.
\newblock {\em Categories for the Working Mathematician}.
\newblock Number~5 in Graduate Texts in Mathematics. Springer-Verlag, 1971.
\newblock (second edition 1997).

\bibitem{PavlovicD:CSR08}
Dusko Pavlovic.
\newblock Network as a computer: ranking paths to find flows.
\newblock In Alexander Razborov and Anatol Slissenko, editors, {\em Proceedings
  of CSR 2008}, volume 5010 of {\em Lecture Notes in Computer Science}, pages
  384--397. Springer Verlag, 2008.
\newblock arxiv.org:0802.1306.

\bibitem{PavlovicD:QI08}
Dusko Pavlovic.
\newblock On quantum statistics in data analysis.
\newblock In Peter Bruza, editor, {\em Quantum Interaction 2008}. AAAI, 2008.
\newblock arxiv.org:0802.1296.

\bibitem{PavlovicD:FAST10}
Dusko Pavlovic.
\newblock Quantifying and qualifying trust: Spectral decomposition of trust
  networks.
\newblock In Pierpaolo Degano, Sandro Etalle, and Joshua Guttman, editors, {\em
  Proceedings of FAST 2010}, volume 6561 of {\em Lecture Notes in Computer
  Science}, pages 1--17. Springer Verlag, 2011.
\newblock arxiv.org:1011.5696.

\bibitem{Poelmans:survey}
Jonas Poelmans, Paul Elzinga, Stijn Viaene, and Guido Dedene.
\newblock {Formal Concept Analysis} in knowledge discovery: {A} survey.
\newblock In Madalina Croitoru, S{\'e}bastien Ferr{\'e}, and Dickson Lukose,
  editors, {\em ICCS}, volume 6208 of {\em Lecture Notes in Computer Science},
  pages 139--153. Springer, 2010.

\bibitem{WagnerK}
Kim~Ritter Wagner.
\newblock Liminf convergence in omega-categories.
\newblock {\em Theor. Comput. Sci.}, 184(1-2):61--104, 1997.

\bibitem{WilleR:FCA}
Rudolf Wille.
\newblock Restructuring lattice theory: an approach based on hierarchies of
  concepts.
\newblock In Ivan Rival, editor, {\em Ordered Sets}, pages 445--470. Dan
  Reidel, Dordrecht, 1982.

\bibitem{Wilson:quasi-metric}
W.A. Wilson.
\newblock On quasi-metric spaces.
\newblock {\em Amer. J. Math.}, 52(3):675--684, 1931.

\bibitem{Kim:pseudo-quasi-metric}
Yong woon Kim.
\newblock Pseudo quasi metric spaces.
\newblock {\em Proc. Japan Acad.}, 10:1009--10012, 1968.

\end{thebibliography}

\end{document}